\pdfoutput=1

\documentclass[11pt]{article}
\usepackage[]{ACL2023}

\usepackage{times}
\usepackage{latexsym}

\usepackage[T1]{fontenc}

\usepackage[utf8]{inputenc}
\usepackage{amsmath}
\usepackage{microtype}
\usepackage{comment}

\usepackage{inconsolata}
\usepackage[export]{adjustbox}

\usepackage{booktabs}
\usepackage[nameinlink]{cleveref}
\usepackage{graphicx}

\title{DecoderLens: Layerwise Interpretation of Encoder-Decoder Transformers}

\author{Anna Langedijk$^1$ \hspace{0.18cm} Hosein Mohebbi$^2$\hspace{0.18cm} \hspace{0.18cm} Gabriele Sarti$^3$ \hspace{0.18cm} Willem Zuidema$^1$ \hspace{0.18cm} Jaap Jumelet$^1$\\[10pt]$^1$ILLC, University of Amsterdam~~$^2$CSAI, Tilburg University~~$^3$CLCG, University of Groningen \\{\small\texttt{annalangedijk@gmail.com}~~~~~\texttt{h.mohebbi@tilburguniversity.edu}}\\ {\small\texttt{g.sarti@rug.nl}~~~~~\texttt{\{w.h.zuidema, j.w.d.jumelet\}@uva.nl}}}
\AfterPreamble{\hypersetup{
  pdfauthor={Anna Langedijk, Hosein Mohebbi, Gabriele Sarti, Willem Zuidema, Jaap Jumelet},
  pdftitle={DecoderLens: Layerwise Interpretation of Encoder-Decoder Transformers}
}}
\begin{document}
\maketitle

\begin{abstract}

In recent years, several interpretability methods have been proposed to interpret the inner workings of Transformer models at different levels of precision and complexity.
In this work, we propose a simple but effective technique to analyze encoder-decoder Transformers. 
Our method, which we name \mbox{DecoderLens}, allows the decoder to cross-attend representations of intermediate encoder activations instead of using the default final encoder output.
The method thus maps uninterpretable intermediate vector representations to human-interpretable sequences of words or symbols, shedding new light on the information flow in this popular but understudied class of models.
We apply DecoderLens to question answering, logical reasoning, speech recognition and machine translation models, finding that simpler subtasks are solved with high precision by low and intermediate encoder layers.

\end{abstract}

\section{Introduction}
Many methods for interpreting the internal states of neural language models -- and in particular Transformer-based models -- have been proposed in the last few years \cite[for a review, see][]{lyu2024towards}.  
Such methods operate at many different levels of granularity, ranging from model-agnostic attribution methods that treat models as black-boxes, to probing methods that assess whether specific information is decodable from model representations, to fine-grained techniques aiming to \emph{causally} link highly localized circuits to model behavior.
These latter techniques (often referred to as `mechanistic interpretability', \citealp{elhage2021mathematical}, or `causal abstractions', \citealp{geiger_causalabstractionsneural_2021}) are often strongly tied to model-specific components, and are likely to provide more faithful insight into how these models operate.

\begin{figure}[t]
    \centering
    \includegraphics[width=\columnwidth]{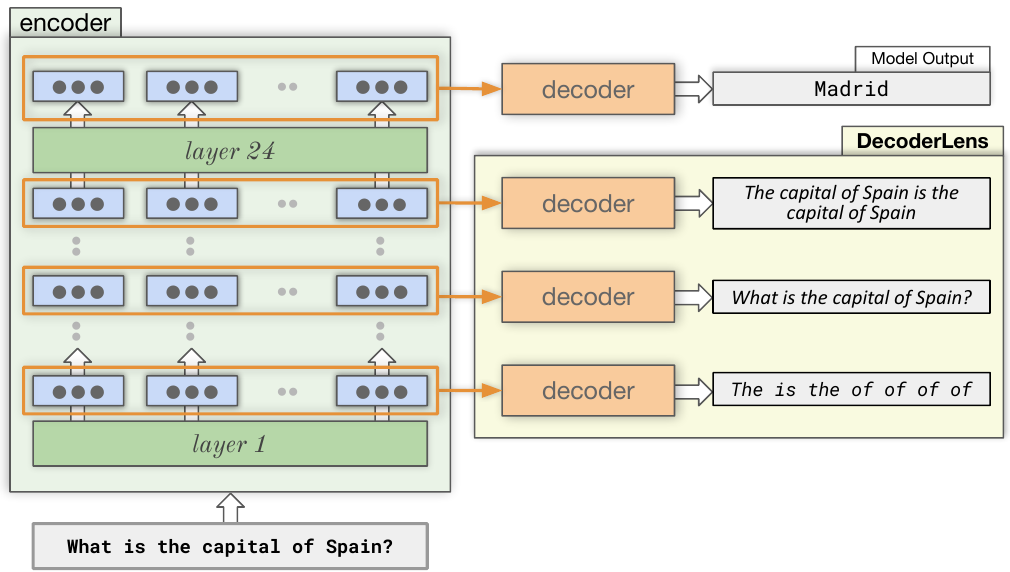}
    \caption{Schematic overview of the DecoderLens. By using the decoder to cross-attend intermediate encoder activation, we can gain qualitative insights into how representations evolve across encoder layers.}
    \label{fig:diagram}
    \vspace{-10pt}
\end{figure} 

In this paper, we propose \emph{DecoderLens}, a method aimed at exploiting the decoder module of encoder-decoder Transformers as a ``lens'' to explain the evolution of representations throughout model layers in these model architectures.
Our method is directly inspired by the LogitLens~\citep{logitlens}, which leverages the \textit{residual stream}\footnote{The sequence of residual connections propagating input information from token embeddings to final layers.} present in Transformer architectures. 
The LogitLens, however, is defined only for decoder-only Transformers, 
and is unable to explain how representations evolve in the encoder of encoder-decoder models.

Concretely, DecoderLens forces the decoder module of an encoder-decoder model to cross-attend intermediate encoder activations. As a consequence, its  generations can be seen as sequences of vocabulary projections depending only on partially-formed source-side representations. 
Such adaptation is necessary as LogitLens requires the presence of a residual stream, which is not found between encoder and decoder modules.
Contrary to common probing methods, DecoderLens operates without any additional training, letting the model ``explain itself'' by producing natural generations in a human-interpretable vocabulary space. \autoref{fig:diagram} provides a graphical overview of our approach.

We evaluate DecoderLens empirically on a wide range of tasks, models, and domains.
First, we demonstrate how representations evolve in Flan-T5 \citep{flanT5} by prompting the model to predict country capitals. %
Next, we conduct an experiment in a more controlled domain, examining how Transformers are able to resolve variable assignment in propositional logic.
{The restricted output space for this task allows us to closely inspect the kinds of solutions intermediate layers produce.}
Finally, we apply the DecoderLens to two common applications of encoder-decoder models: 
neural machine translation~\citep{team-2022-nllb} and speech-to-text transcription and translation~\citep{Radford2022RobustSR}.

We find that intermediate outputs can be useful to find hypotheses about the strategies a model uses for solving (sub)tasks.
One surprising finding, for example, is that Flan-T5 encodes geographical information \textit{better} in intermediate layers than in the top layer. 
Additionally, our findings show that the middle encoder layers approximate correct transcriptions and translations well for models such as Whisper and NLLB.
Experiments from both logic and machine translation show that earlier layers sometimes output local approximations to their respective tasks.
The DecoderLens thus provides a useful tool that can be used in combination with other interpretability methods to gain a more complete insight into the inner workings of deep encoder-decoder language models.

\section{Related Work}
The current state of interpretability methods can be categorized by the different levels of granularity at which they explain model behavior.
At the coarsest level, model-agnostic methods such as feature attributions \citep[e.g.,][]{sundararajan2017axiomatic,lundberg2017unified} focus on explaining model output in terms of the most important input features.
A major concern with  this line of work is the \textit{faithfulness} of a method: whether the attributions the method  produces in fact correspond to the true, underlying causes of the model's output. 
The strong disagreement between different attribution methods raises doubts that the faithfulness requirement is met in practice 
\citep{DBLP:conf/acl/JacoviG20, DBLP:conf/hhai/NeelySBL22,lyu2024towards}. 

In response to these concerns, a novel line of work that has received increasing attention in recent years attempts to explain models at a more fine-grained level, leveraging knowledge about a model's inner workings based on specific components \citep[e.g.,][]{elhage2021mathematical, NEURIPS2022_6f1d43d5, mohebbi-etal-2023-quantifying, wang2023interpretability}.

\paragraph{Interpreting Language Models in Vocabulary Space} A common way of studying Transformers in this line of work is to take advantage of the {\it residual stream}.
In this view, each layer can be seen as adding or removing information by reading from or writing to the hidden states in the residual stream \citep{elhage2021mathematical}.
LogitLens \citep{logitlens} uses this idea by directly applying the unembedding operation to the middle layers of GPT to obtain a logit distribution for every intermediate layer.
As the method 
projects into the output (logit) space, it can provide interpretable insights about which information arises in which layers. This is similar to the projections into vocabulary space used to verbalize probing methods in earlier work \citep{DBLP:conf/naacl/SaphraL19, DBLP:conf/acl/JumeletDSHS21}.

\citet{merullo2023language} use the Logit Lens to identify different generic stages of processing throughout GPT's layers in a Question Answering task. 
\citet{halawi_overthinkingtruthunderstanding_2023} instead use the Logit Lens to study {\it overthinking}, identifying critical layers in which the logit distribution suddenly shifts to an incorrect prediction.
\citet{geva-etal-2022-transformer} use the idea of the residual stream to study what kind of {\it updates} happen in each feed-forward layer, by analyzing the differences in logit outputs between layers. 
The updates are in vocabulary space, making them easily interpretable to humans.
Similarly, \citet{dar-etal-2023-analyzing} also project other Transformer components into vocabulary space, such as its attention weights, and find that these can encode coherent concepts and relations.
\citet{tunedlens} present the Tuned Lens, extending the Logit Lens with an optimized, affine transformation before the unembedding operation, and report that it produces more reliable and predictive results. Finally, \citet{ghandeharioun2024patchscope} present a general framework for information lenses called \texttt{Patchscopes}, and show that auxiliary models can be tuned to act as expressive vocabulary projections.

\paragraph{Early Exiting in Language Models} Early exiting enables models to make early predictions by skipping subsequent layers once the model reaches sufficient confidence, improving model efficiency by speeding up inference. 
This is usually achieved by training intermediate classifiers on top of each encoder layer in encoder-only models \citep{liu-etal-2020-fastbert, NEURIPS2020_d4dd111a, schwartz-etal-2020-right, liao-etal-2021-global, xin-etal-2020-deebert, xin-etal-2021-berxit}, or by training intermediate unembedding heads for each decoder layer in decoder-only or encoder-decoder models \citep{schuster2022confident}. 
\citet{pal-etal-2023-future} find that early-exiting from intermediate token representations can produce accurate next token predictions for several generation steps ahead, exploiting the parallel nature of Transformers outputs.  
Similar to early exiting is the concept of \textit{encoder layer fusion}, in which a decoder can cross-attend to all encoder layers instead of the final one.
This allows the decoder to use surface-level representations from early layers in addition to abstract, highly contextualized representations from later layers, which can improve the final performance \citep{dou-etal-2018-exploiting, liu2020understanding, Feng2021EncoderFN, georges-gabriel-charpentier-samuel-2023-layers}. 

\begin{figure*}[!th]
    \centering
    \includegraphics[width=.72\columnwidth, trim={0 0 4cm 0}, clip]{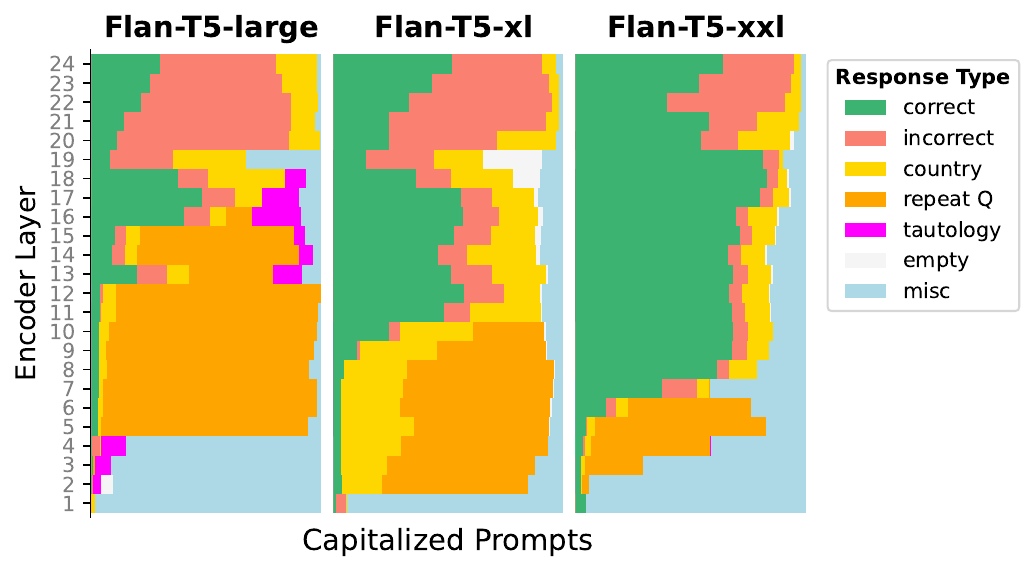}
    \hfill
    \includegraphics[width=.89\columnwidth, trim={0.8cm 0 0 0}, clip]{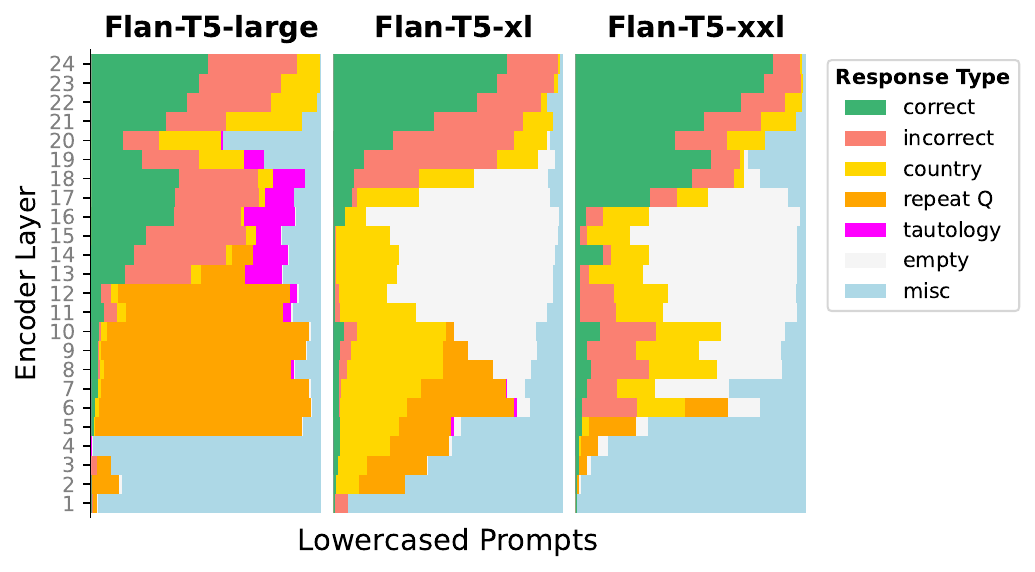}
    \hfill
    \includegraphics[width=.38\columnwidth, trim={0 0 0.8cm 0}, clip]{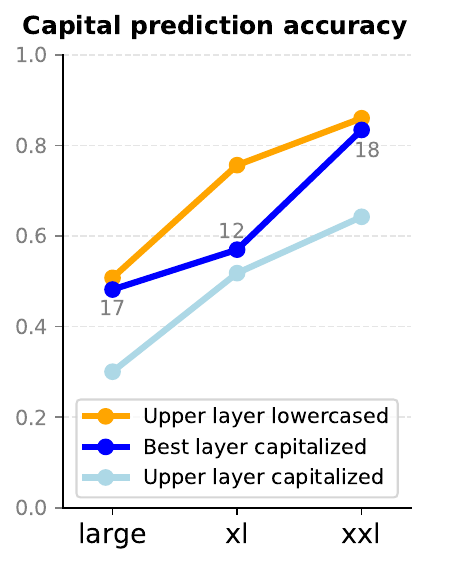}
    \caption{
    Distribution of response types for three Flan-T5 models on the country capital prediction task.
    Each row indicates the encoder layer that was used for the DecoderLens.
    Capital prediction accuracy denotes the model performance on the task for the two prompt types, including the best performing layers for the capitalized prompts.
    }
    \label{fig:t5}
\end{figure*}
\section{DecoderLens}
The DecoderLens approach is inspired by the LogitLens method of \citet{logitlens}. 
The main intuition behind this method is that the residual stream in Transformer decoder-only models forces representations across layers to gradually converge towards the final representation, iteratively refining its guess \citep{DBLP:conf/iclr/JastrzebskiABVC18, DBLP:conf/iclr/DehghaniGVUK19}.
This gradual change allows us to inspect how model predictions change across layers by directly applying the final \textit{unembedding} transformation to intermediate hidden states.

For encoder-decoder models, the LogitLens can only be applied to the decoder component since there is no residual stream between encoder and decoder modules.
To investigate how representations in the encoder evolve across layers, we therefore introduce the \textbf{DecoderLens}, which leverages the \textit{entire decoder} to verbalize the knowledge captured by intermediate encoder layers.
This is achieved by early exiting the encoder at earlier layers, and using the resulting representations for the decoder cross-attention operation.
The DecoderLens allows for richer insights than the LogitLens, enabling the generation of full outputs from intermediate encoder states. It also may help mitigate out-of-distribution issues that can arise from using a single vocabulary projection \citep[e.g.][]{tunedlens, yomdin-etal-2023-jump}.
The model outputs plausible strings that adhere to the original training objective, allowing us to see {\it how} the task is progressively addressed throughout encoder layers. 

We define the DecoderLens as follows.
For an encoder-decoder model $\mathcal{M}$ with $n$ layers, the output of the decoder is normally generated based on the top-layer representations of the encoder, combined with a decoding algorithm (e.g. beam search).
Often, the encoder layers are first passed through a non-linear operation, such as layer normalization \citep{DBLP:journals/corr/BaKH16}.
The DecoderLens operates similarly, by first passing the $i^{\textup{th}}$ encoder layer through the non-linear operation $f$, and then feeding it as input to the decoder:
\begin{align*}
    \mathcal{M}(\mathbf{w}) &= Dec\left(f(Enc(\mathbf{w})_n)\right) \\
    \textit{DecoderLens}(\mathbf{w}, i) &= Dec\left(f(Enc(\mathbf{w})_i)\right)
\end{align*}
In the following sections, we investigate the effectiveness of DecoderLens by applying it to a variety of tasks, models, and domains. 
\section{Factual Trivia QA}
We first apply the DecoderLens to investigate the factual knowledge of a instruction-tuned encoder-decoder LM, Flan-T5 \citep{flanT5}\footnote{All pre-trained models in the paper were evaluated via the \texttt{transformers} library~\citep{wolf-etal-2020-transformers}}.
As a case study, we consider country capital prediction, using prompts of the form ``\textit{What is the capital of X?}'' and testing encoder layers' ability to produce correct outputs for all 193 United Nations member states.
We evaluate Flan-T5 models of three sizes (\textit{large}, \textit{xl}, \textit{xxl}, with 0.78B, 3.0B and 11.3B parameters respectively), containing the same number of layers (24) and hidden state size (1024), but differing in the feed-forward layer size and the number of attention heads \citep{DBLP:journals/jmlr/RaffelSRLNMZLL20}.

\paragraph{Evaluation}
To investigate the types of responses generated by the DecoderLens, we categorize model answers as follows: 1) correct response, based on a full string match, 2) incorrect response in the form of a different city name, 3) country name itself, 4) repetition of the question, 5) tautologies (\textit{The capital of X is the capital of X}), 6) empty responses containing no alphanumeric characters, and 7) a miscellaneous category for anything that doesn't fall under these previous six categories.
These categories were defined after a manual inspection of the DecoderLens results: some examples of intermediate outputs can be seen in \autoref{tab:t5_qualitative_examples}. We conduct the experiment on lowercased and capitalized prompts to test the robustness of the model to minimal variations in the provided inputs.

\begin{table}[h]
    \footnotesize
    \begin{tabular}{l|l}\toprule
         Layer  & Output \\\midrule
         \bf L0 & What\\ 
         \bf L3 & What is the capital of Colombia? \\
         \bf L8 & What is the capital of Colombia? \\
         \bf L12 & \it The capital of Colombia is Bogotá. \\
         \bf L16 & Colombians are a very friendly people. \\
         \bf L19 & Buenos Aires \\ 
         \bf L21 & colombia \\
         \bf L24 & \it bogota \\
         \bottomrule
    \end{tabular}
    \caption{DecoderLens predictions for ``What is the capital of Colombia?" for various Flan-T5-\textit{xl} encoder layers. Correct outputs are {\it italicized}. }
    \label{tab:t5_qualitative_examples}
    \vspace{-10pt}
\end{table}

\paragraph{Results}

We present the results for the experiment in \autoref{fig:t5}.
Capitalized and lowercased prompts yield considerably different patterns across layers.
For capitalized prompts, we surprisingly find that all models yield \textit{better performances} for intermediate layers compared to the canonical top layer of the model.
For lowercased prompts, on the other hand, the top layer always yields the highest accuracy of all layers.
The difference between the capitalized and lowercase prompts suggests that geographical knowledge is stored in different locations based on capitalization.
We speculate that this might be due to the more frequent splitting of lowercased country names into multiple subtokens (188 out of 193 countries) compared to capitalized country names (only 87 out of 193 countries, including multi-word country names).
Hence, country names split into multiple subtokens need to be compositionally combined by the model before retrieving their capital from encoded representations.

Finally, we note that Flan-T5-\textit{large} has a long phase in which the DecoderLens results in a repetition of the original query prompt.
In the \textit{xl} model this occurs in lower layers, alongside repetitions of the country name itself, while the \textit{xxl} model is less prone to these patterns, producing correct results much earlier for the capitalized case.

\section{Propositional Logic}
Results from the previous section indicate that DecoderLens can be useful for identifying the layers in which factual information arises and can be readily decoded in general pre-trained language models.

In this section, we go one step further and apply DecoderLens to a model exclusively trained on a downstream task.
We believe it is advantageous to test novel interpretability methods on models that are trained to solve a simple, unambiguous task within a carefully controlled setup \citep[][]{hupkes2018visualisation, hao-2020-evaluating, jumelet-zuidema-2023-feature, nanda_progressmeasuresgrokking_2023,nanda-etal-2023-emergent}.

We apply the DecoderLens to a small Transformer model that is trained from scratch on a synthetic (but non-trivial) task: predicting variable assignments given a logical formula.

\paragraph{Task}
We study an encoder-decoder model that is specifically trained on propositional logic, based on the setup of \citet{hahn_teachingtemporallogics_2021}. 
The model is trained to output a partial satisfying {\it assignment} given a satisfiable formula in propositional logic. 
These inputs consist of logical operators ({\sc not}/$\neg$/\texttt{!}, {\sc and}/$\wedge$/\texttt{\&}, {\sc or}/$\vee$/\texttt{|}, {\sc iff}/$\leftrightarrow$ and {\sc xor}/$\oplus$) and at most five propositional variables. 

\autoref{tab:logic_data_examples} lists a few examples.

\begin{table}[h]
  \centering\small
  \begin{tabular}[h]{l|l|l}
    \toprule
    Formula & Input & Output \\\midrule
    $ \neg a \wedge ( b \vee c )$ & \texttt{\& !\ a | b c}& a 0 b 1\\
    $ a \oplus \neg e $ & \texttt{xor a !\ e} &a 1 e 1\\ \bottomrule
  \end{tabular}
  \caption{Example datapoints for two formulas. Inputs use prefix notation to avoid the use of parentheses. The first assignment is {\it partial}: the value of \texttt{c}  could be either \texttt{0} or \texttt{1}, and may therefore be omitted.}\label{tab:logic_data_examples} 
\end{table}

The models are trained in a standard sequence-to-sequence setup using teacher forcing and only have access to a single correct output, even when several partial assignments would be semantically correct.
Nevertheless, this limited setup seems sufficient to teach these models the semantics of propositional logic \citep{hahn_teachingtemporallogics_2021}.
At test time, the models are able to output novel assignments to unseen formulas with 93\% accuracy.

\paragraph{Experimental setup}
We test encoder-decoder models using the standard transformer architecture. Encoder and decoder modules have each six layers, with hidden sizes of 128 and 64 respectively.
Models are trained for 128 epochs on the {\it PropRandom35} training set of \citet{hahn_teachingtemporallogics_2021}, which consists of 800k randomly generated formulas containing at most 35 symbols.
The ground truth output assignments are generated by a symbolic SAT solver using \texttt{pyaiger} \citep{pyaiger}.
We train three different model seeds and aggregate the results.

\subsection{Evaluation on Controlled Data}\label{sec:logic:data}
We apply the DecoderLens to 1) randomly generated data and 2) handcrafted formulas using templates of varying difficulty. We hypothesize that easier formulas can be solved in earlier layers.

First, we evaluate on the {\it PropRandom35} validation set of 200k sentences, and an additional dataset of 200k short sentences with a maximum length of 12, {\it PropRandom12}.\footnote{These shorter sentences are easier to automatically group into varying levels of difficulty.}
Second, to gain more insight into the types of formulas layers can solve, we generate a dataset according to four templates:
\begin{itemize}
\setlength\itemsep{0.1em}
    \item[\textsc{t1}.] \textbf{Simple conjunction:} formulas in the form of $l_1 \wedge l_2 \wedge l_3 \wedge l_4$, where $l_n$ is a propositional literal ($p$ or $\neg p$). These formulas can be solved ``locally", simply by reading the truth value from each variable separately. 
    \item[\textsc{t2}.] \textbf{Local XOR:} formulas in the form of $(l_1 \oplus l_2) \wedge (l_3 \oplus l_4)$, where all literals are distinct. Variables interact with their siblings via $\oplus$, but the two parts of the formula can be solved independent of one another.
    \item[\textsc{t3}.] \textbf{Non-local XOR:} formulas in the form of $(l_1 \oplus l_2) \wedge (l_3 \oplus l_4)$, where $l_2$ and $l_3$ contain the same variable. The two parts cannot necessarily be solved independently.
    \item[\textsc{t4}.] \textbf{Non-local CNF:} formulas in the form of $(p_1 \vee \neg p_2) \wedge (p_2 \vee \neg p_3) \wedge (p_3 \vee \neg p_1)$, containing  dependencies between the clauses: this means the formulas cannot be solved locally.
\end{itemize}
For each template, we generate all possible non-trivial variable combinations, for multiple orderings of the subformulas. We filter out formulas that are not solved by the models. The total size of the template dataset is 30k.\footnote{All datasets used for evaluation are available at \href{https://github.com/annaproxy/decoderlens-data}{github.com/annaproxy/decoderlens-data}}

\begin{figure}
    \centering
   \includegraphics[width=.9\linewidth]{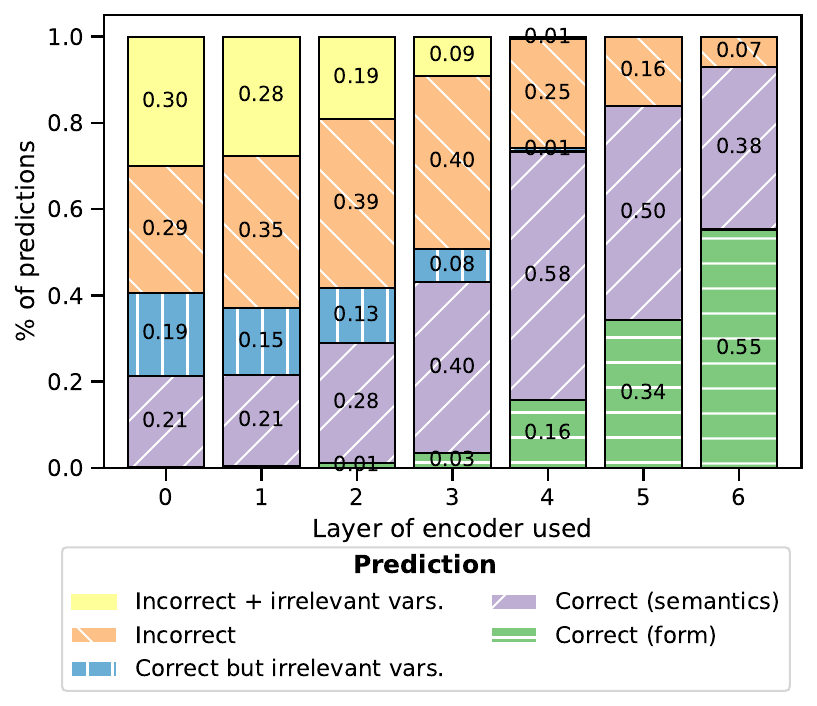}
    \caption{Performance of DecoderLens using intermediate encoder layers on the PropRandom35 validation set. Layer 0 denotes the embedding layer. 
    The category {\it correct (semantics)} denotes outputs are correct, but deviate from ground truth sequences. 
    All outputs in the {\it correct (syntax)} category are also semantically correct.
    We define variables as \textit{irrelevant} when they did not occur in the input, but appear in the prediction.}
    \label{fig:logic_performance}
    \vspace*{-0.8em}
\end{figure}

\paragraph{Results} 
We evaluate the DecoderLens on the validation set of {\it PropRandom35}: the results are shown in \autoref{fig:logic_performance}. We manually inspect some intermediate model outputs  (\autoref{tab:logic_qualitative_examples} lists some examples).

We observe that nearly all incorrect outputs are still in the correct format, although many contain irrelevant variables that do not occur in the input formula.
This suggests a learned division of duties between the encoder and decoder, with the decoder being completely in charge of formatting and variable ordering.\footnote{Even when random noise is passed to the decoder, it still outputs variables and their truth values in the correct order.}
Note that there are a limited number of possible correctly formatted outputs (242 in total), of which, on average, 29\% are semantically correct. 
The total semantic accuracy of the embedding layer and the first two layers is below 29\%, meaning they do not perform better than random chance.
Moreover, we find that initial layers often produce irrelevant variables, suggesting that their representations are misaligned with the final layer representations to an extent that makes them uninformative for the decoder.

Layers three and four prune these irrelevant variables and perform well above chance level. Examples of formulas that are already solved by these layers are the first two formulas in \autoref{tab:logic_qualitative_examples}.

We observe that another function of the final two layers is to prune contingent variables, refining an already correct solution. 
E.g., in the first example in \autoref{tab:logic_qualitative_examples}, layer five refines the layer four solution by removing the unnecessary ``c 1".
Around 20\% of outputs of layers 5 and 6 are strict sub-outputs of the previous layer, removing 1.3 variables from the previous output on average.
In a small number of cases (2.6\%), layer five outputs a correct assignment but layer six does not: this could be seen as the model {\it overthinking} the output \citep{halawi_overthinkingtruthunderstanding_2023}. 
Only a minority of these cases (20\% of the 2.6\%) are due to layer six pruning a necessary variable.

The examples in \autoref{tab:logic_qualitative_examples} also demonstrate that solutions are more {\it local} in earlier layers. 
For instance, in the second example, layer three assigns {\it false} to both \texttt{a} and \texttt{d}, as they both occur negated in the sentence. 
The operator \textsc{xor}, which requires communication between the two variables, is not taken in consideration yet.

\begin{table}[h]
    \centering\small
    \begin{tabular}{l|lll}\toprule
         Layer  & $\neg b \wedge (c \vee a)$ & $\neg d \oplus \neg a$ & $b \oplus (b \wedge a)$ \\\midrule
         \bf L0 & a 0 b 1 e 0    & a 1 b 1 c 1 e 1    & a 1 b 1 c 0 e 0     \\
         \bf L1 & a 1 b 1 e 0    & a 1 b 1 d 1 e 1    & a 1 b 1 e 1         \\
         \bf L2 & \it a 1 b 0 c 1 & a 0 d 0 e 0    & a 1 b 1 c 0 e 1 	   \\
         \bf L3 & \it a 1 b 0 c 1 & a 0 b 0 d 0    & a 1 b 1 e 0         \\
         \bf L4 & \it a 1 b 0 c 1 & \it a 0 d 1    & a 1 b 1             \\
         \bf L5 & \it a 1 b 0     & \it a 0 d 1    & \it a 0 b 1 	         \\
         \bf L6 & \it b 0 c 1     & \it a 0 d 1    & \it a 0 b 1 	         \\\bottomrule
    \end{tabular}
    \caption{DecoderLens predictions on three simple logical formulas across encoder layers. L0: embedding layer. Semantically correct outputs are {\it italicized}. }
    \label{tab:logic_qualitative_examples}
    \vspace{-10pt}
\end{table}

\subsection{Locality of Intermediate Outputs}
\label{sec:logic:locality}

To further investigate the locality of model outputs across encoder layers, we apply the model to multiple sets of sentences based around the \textsc{xor}-operator and its logical opposite, \textsc{iff}.
We group the short formulas from {\it PropRandom12} into three categories: one where neither operator is present, one where either operator is present but is not the direct parent of another \textsc{xor}/\textsc{iff} (e.g. $(a \leftrightarrow b) \wedge (b \oplus c)$), and one having at least one nested instance of these two operators (e.g. $(a \oplus b) \leftrightarrow (c \wedge b)$).
These patterns can be indicators of the formula's difficulty, but random formulas are not guaranteed to be (non)local.
We therefore also analyze the performance of earlier layers on the handcrafted sentences described in \S\ref{sec:logic:data}.

\begin{figure}[t]
    \centering
    \includegraphics[width=\linewidth]{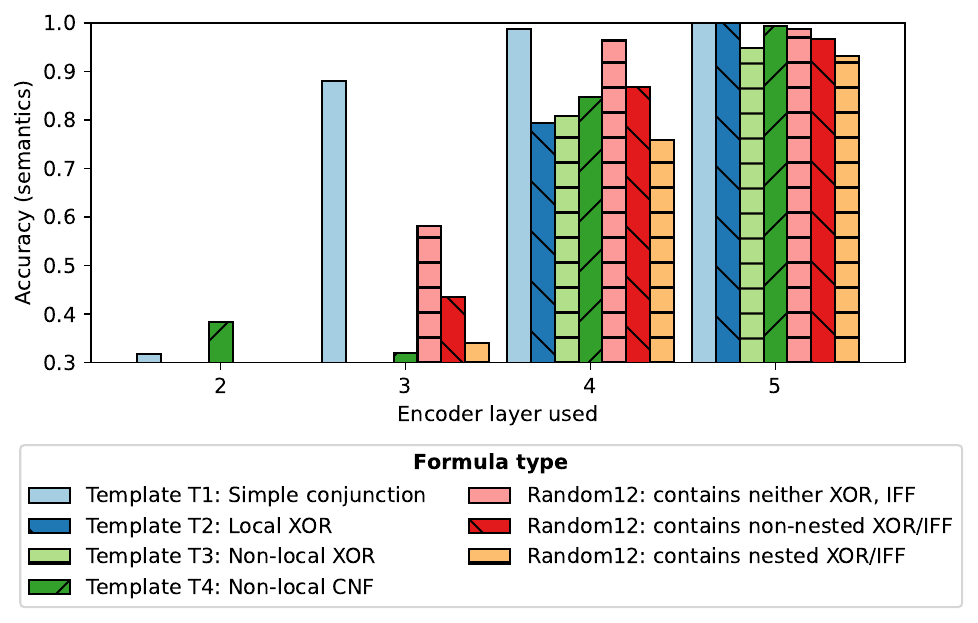}
    \caption{Performance on different kinds of formulas for the middle encoder layers.}
    \label{fig:logic_groups}
\end{figure}

\paragraph{Results} DecoderLens results for the formula types detailed above are presented in \autoref{fig:logic_groups}. 
We observe large jumps in performance across layers for the different sets of formulas. 
In particular, we note that simple conjunctions (pattern \textsc{t1}) can already be solved in layer three. However, the same layer cannot solve formulas including \textsc{xor}. Instead, the layer outputs a {\it local} solution as in example 2 in \autoref{tab:logic_qualitative_examples}, by simply assigning {\it 0} to each variable that occurs in the negated inputs, and {\it 1} in the non-negated ones.

Overall, a local solution is produced for at least one of the subformulas in 87\% of cases, and for both formulas in 53\% of cases.
Other layers output local solutions as a much lower rate: more details can be seen in \autoref{fig:logic_appendix} in \autoref{sec:logic:appendix}.
Layer four sees the largest improvement for all other types of formulas, but still lags behind in solving non-local formulas, especially those containing nested \textsc{xor} or \textsc{iff}-operators. 

These results supports the intuition that the model {\it gradually refines} its prediction by contextualizing its representations: first, variables collect {\it local} information about their possible truth value. 
These variables can only exchange information with other variables in the later layers to reach a coherent solution.

\section{Machine Translation}
\label{sec:mt}

We apply DecoderLens to NLLB-600M~\citep{team-2022-nllb}, a state-of-the-art multilingual model trained in over 200 languages, to quantify encoder influence on translation quality and properties. 
We use 1012 sentences from the dev/test split of Flores-101~\citep{goyal-etal-2022-flores}, using English $\leftrightarrow$ \{Italian, French, Dutch\} as high-resource pairs and English $\leftrightarrow$ \{Xhosa, Zulu\} as low-resource pairs to evaluate differences in intermediate encoder layers' performances for these two settings.

\begin{figure}
    \centering
    \includegraphics[width=\linewidth]{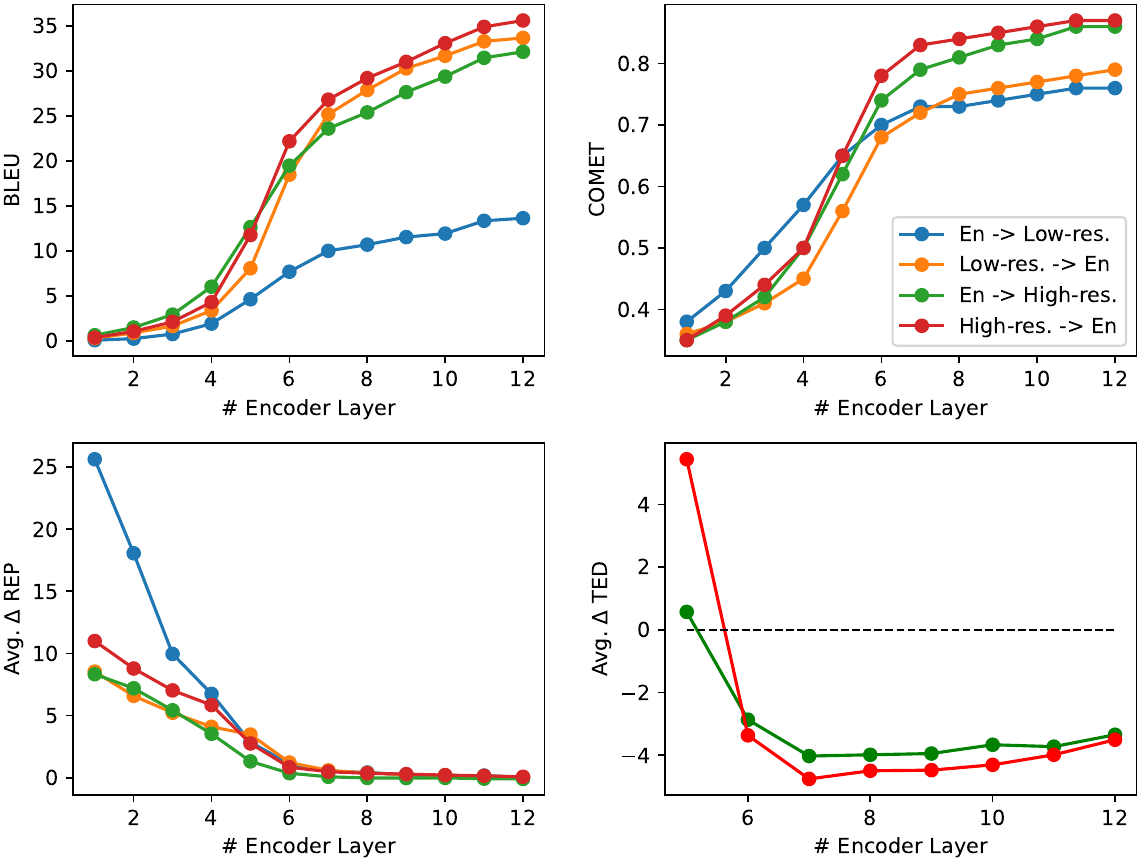}
    \caption{Performance of NLLB across encoder layers. Scores are averaged across into-English (XX $\rightarrow$ EN) and from-English directions (EN $\rightarrow$ XX) for low-resource and high-resource languages.}
    \label{fig:mt_results}
    \vspace{-10pt}
\end{figure}

\paragraph{Metrics} We evaluate the translation quality of DecoderLens outputs using BLEU and COMET~\citep{papineni-etal-2002-bleu,rei-etal-2022-comet}. 
Moreover, we define two ad-hoc metrics to estimate additional output properties.
To quantify \textit{repetition} we compute the difference in counts between most common tokens in the output and reference translation ($\Delta$\textsc{rep}).
To measure \textit{syntactic reordering}, we compute tree edit distance (\textsc{ted}) between source and target syntax trees, for both the output (\textsc{ted}$_\textsc{out}$) and reference translations (\textsc{ted}$_\textsc{ref}$). We then take their difference: $\Delta$\textsc{ted} $=$ \textsc{ted}$_\textsc{out} - $\textsc{ted}$_\textsc{ref}$. 
Positive scores for this metric reflect more syntactic reordering in the output compared to the reference translation. {A negative $\Delta$\textsc{ted} indicates that the model translation adheres more closely to the source sentence word ordering than the reference translation does.}
We limit our \textsc{ted} evaluation to layers with BLEU > 10 and high-resource pairs, using the Stanza, FastAlign and ASTrED libraries ~\citep{qi-etal-2020-stanza,dyer-etal-2013-simple,vanroy2021metrics} for parsing, alignment and \textsc{ted} computations respectively.

\begin{table}
\centering
\small
\adjustbox{max width=\linewidth}{
\begin{tabular}{l}
\toprule
\textbf{Source:} In late 2017, Siminoff appeared on shopping television channel QVC.\\
\textbf{Reference:} Fin 2017, Siminoff est apparu sur la chaîne de télé-achat QVC.\\
\midrule
\textbf{L1:} Dans la télévision, il est possible de faire une pause dans la conversation. \\
\textbf{L2:}  Dans le cas de la télévision, il est possible de faire une demande de renseignement. \\
\textbf{L3:}  En 2017, le téléviseur a été mis au défi de la télévision. \\
\textbf{L4:}  En 2017, le canal de télévision de la télévision a été \underline{mis en vente}. \\
\textbf{L5:}  En 2017, Siminoff est apparu sur la chaîne de télévision QVC. \\
\textbf{L6:}  En 2017, Siminoff est apparu sur la chaîne de télévision QVC. \\
\textbf{L7:}  En 2017, Siminoff est apparu sur la chaîne de télévision de \underline{shopping} QVC. \\
\textbf{L8:}  En 2017, Siminoff est apparu sur la chaîne de télévision de shopping QVC. \\
\textbf{L9:}  En 2017, Siminoff est apparu sur la chaîne de shopping TV QVC. \\
\textbf{L10:} En 2017, Siminoff est apparu sur la chaîne de télévision de shopping QVC. \\
\textbf{L11:} \underline{Fin} 2017, Siminoff est apparu sur la chaîne de télévision de shopping QVC. \\
\textbf{L12:} Fin 2017, Siminoff est apparu sur la chaîne de télévision de shopping QVC. \\
\bottomrule
\end{tabular}
}
\caption{Example DecoderLens translations for an English $\rightarrow$ French sentence of Flores-101.}
\label{tab:mt_example}
\vspace{-10pt}
\end{table}

\paragraph{Quantitative results} \autoref{fig:mt_results} presents the results of our evaluation. We remark a stark difference in quality between translation into low-resource languages and other settings, with performance increasing rapidly halfway through encoder layers only in the latter case. 
All language directions exhibit a large number of repetitions for the first half of the encoder, suggesting that initial encoder layers are generally tasked to model $n$-gram co-occurrences, as also noted by \citet{voita-etal-2021-language} for initial phases of neural MT training. 
Repetitions decline to match reference frequency around models' intermediate layers, coinciding with the largest increase in translation quality. 
Regarding reordering, syntax in translations stabilizes early in the encoder layers: in line with previous findings~\citep{vanroy-2021-syntactic}, outputs show a lower degree of syntactic reordering relative to source texts when compared to human references, providing additional evidence about the locality of intermediate layers' predictions shown in Section~\ref{sec:logic:locality}.
The lack of spikes in translation quality for intermediate encoder layers in low-resource directions using DecoderLens can be connected to the low source context usage shown in~\citet{ferrando-etal-2022-towards}, suggesting that poor intermediate outputs for these directions might be due to the out-of-distribution behavior of the decoder component.

\paragraph{Qualitative results} We manually examine a subset of 50 DecoderLens translations through encoder layers (\autoref{tab:mt_example}, more examples in ~\Cref{app:mt_examples}). 
For high-resource pairs, translations in the first few layers are fluent and contain keywords from the original sentence, but are completely detached from the source {(see for example the L1 output in \autoref{tab:mt_example}, which contains the word ``television" but is otherwise detached from the English source. )}.
Intermediate layers often output examples with incorrect word sense disambiguation (e.g. ``shopping TV channel'' interpreted as ``TV channel being sold'' in L4). Finally, more granular information is often added at later stages (e.g. ``shopping'' added in L7 and ``Fin'' in L11).
\section{Speech-to-Text}

\begin{figure}
    \centering
    \includegraphics[width=\linewidth]{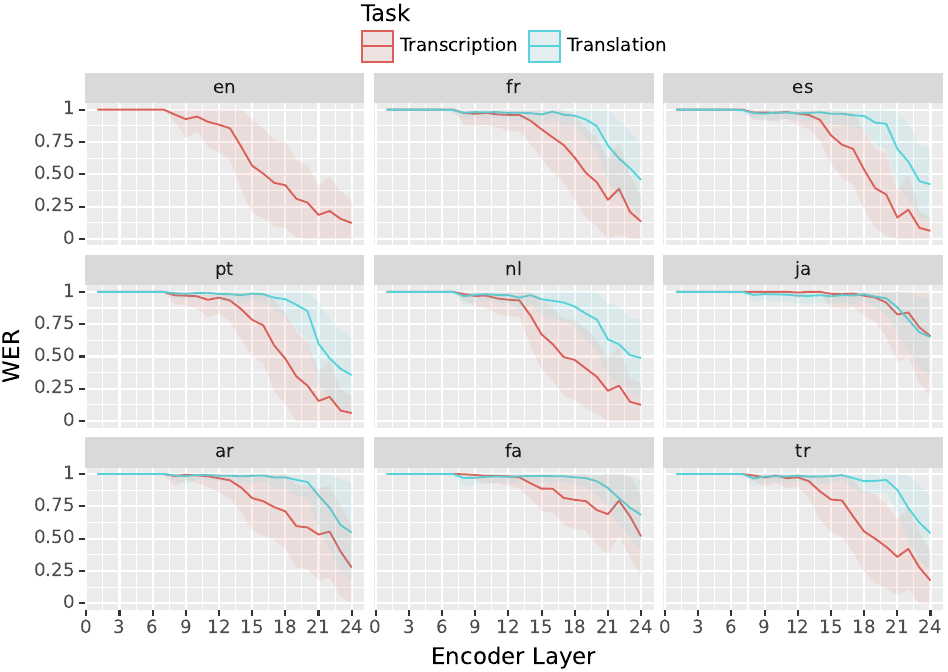}
    \caption{Average Word Error Rate (wer) of Whisper-medium for transcription and translation w.r.t number of encoder layer used at inference. Shaded areas show one standard deviation.}
    \label{fig:speech_wer_medium}
    \vspace{-10pt}
\end{figure}
We next apply DecoderLens to Whisper \citep{Radford2022RobustSR}, a state-of-the-art multilingual speech model trained on a set of supervised audio-to-text tasks, including multilingual speech transcription and speech translation to English. We use Whisper in three different sizes ({\it base}, {\it small}, and {\it medium}) which differ in their number of layers (6, 12, and 24, respectively).

\paragraph{Data}
We use \mbox{CoVoST 2} \citep{wang2020covost}, a multilingual speech-to-text translation dataset based on \mbox{Common Voice} corpus \citep{Ardila2019CommonVA}. 
We sample 100 sentences for nine languages: English (en), French (fr), Spanish (es), Portuguese (pt), Dutch (nl), Japanese (ja), Arabic (ar), Persian (fa), and Turkish (tr).
Since the dataset includes both source and translation references for each utterance, we can inspect Whisper's behavior for both transcription and translation tasks on the same examples, providing an unbiased comparison between the tasks.

\begin{table}
\centering
\small
\adjustbox{max width=\linewidth}{
\begin{tabular}{l}
\toprule
\textbf{Input utterance:} turning off gadgets that are not in use can save a lot of energy\\
\midrule
\textbf{L1-7:}  \\
\textbf{L8:~~} ``of the world'' \\
\textbf{L9:~~} ``tornado'' \\
\textbf{L10:~} ``i am going to talk about the new technology \underline{that} we have'' \\
\textbf{L11:~} ``tornado'' \\
\textbf{L12:~} ``i am going to go ahead and say that i am \underline{a} little bit more of a fan of the channel...'' \\
\textbf{L13:~} ``i am going to go ahead and turn it over to you and i am going to turn it over to you and...'' \\
\textbf{L14:~} ``tony i am glad you \underline{are} here'' \\
\textbf{L15:~} ``\underline{turning off gadgets that} are \underline{not} news \underline{can save} a \underline{lot of energy}'' \\
\textbf{L16:~} ``turning off gadgets that are not news can save a lot of energy'' \\
\textbf{L17:~} ``turning off gadgets that are not news can save a lot of energy'' \\
\textbf{L18:~} ``turning off gadgets that are not news can save a lot of energy'' \\
\textbf{L19:~} ``turning off gadgets that are not news can save a lot of energy'' \\
\textbf{L20:~} ``turning off gadgets that are not used can save a lot of energy'' \\
\textbf{L21:~} ``turning off gadgets that are not \underline{in use} can save a lot of energy'' \\
\textbf{L22:~} ``turning off gadgets that are not in use can save a lot of energy'' \\
\textbf{L23:~} ``turning off gadgets that are not in use can save a lot of energy'' \\
\textbf{L24:~} ``turning off gadgets that are not in use can save a lot of energy'' \\
\bottomrule
\end{tabular}
}
\caption{Whisper-medium intermediate transcription outputs for an English utterance. Words that are correctly generated for the first time are \underline{underlined}.}
\vspace{-10pt}
\label{tab:speech_examplewise_medium}
\end{table}

\begin{figure}[ht]
    \centering
    \includegraphics[width=\linewidth]{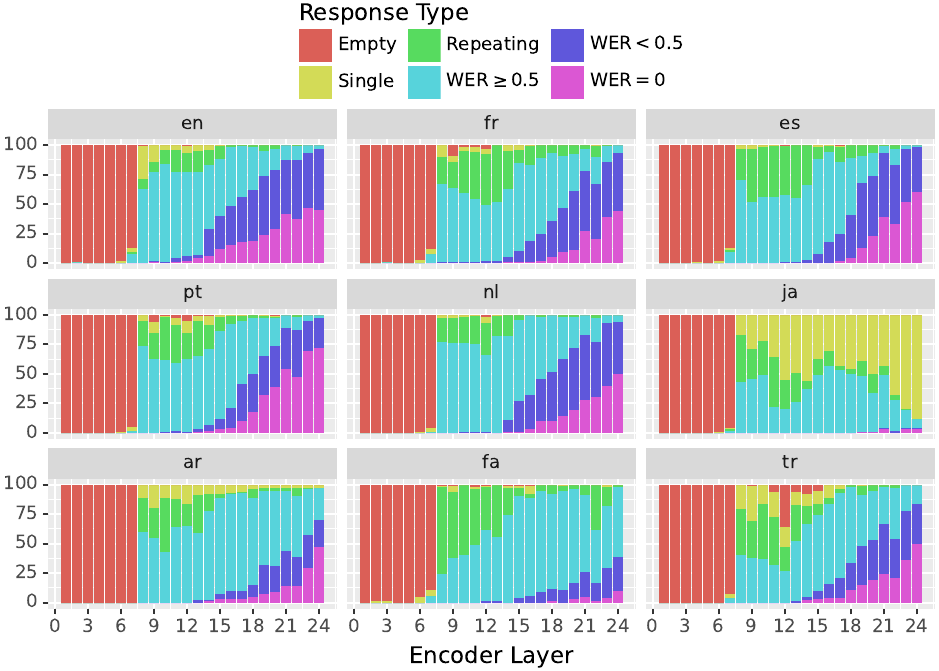}
    \caption{Distribution of Whisper-medium output types when transcribing, for each encoder layer.}
    \label{fig:speech_type_medium}
    \vspace{-10pt}
\end{figure}

\paragraph{Results}
\autoref{fig:speech_wer_medium} shows the overall results of Word Error Rate (WER) across various source languages when applying DecoderLens to Whisper-{\it medium} for transcription and translation tasks. 
While the overall pattern of WER is decreasing, we can discern that fundamental information emerges from the intermediate layers. 
Comparing the trend of WER for transcription and translation, it appears that the essential information required for transcription is encoded in earlier encoder layers compared to translation.\footnote{The same pattern is observed for the other model sizes, reported in \Cref{app:speech_wer}.} \autoref{tab:speech_examplewise_medium} shows a more fine-grained view of the changes in model output transcription.
The first 7 layers of the encoder produce empty outputs, indicating that the information is not yet ready for transcribing. Next, layers 8-11 generate a limited number of irrelevant words (notably, generating single words in layers 9 and 11), while layers 12-13 produce long sequences of repeating irrelevant words. The main part of the true transcription can be constructed starting from layer 15 (with some minor errors; the word `news' is generated instead of `in use' in this example). 
The error in this running example is then corrected in layer 21, and this information is carried to the final encoder layer.  \autoref{fig:speech_type_medium} quantifies this to show that the pattern holds for the majority of examples in all languages. 
This pattern holds for both tasks and different model sizes, except for the early encoder layers of Whisper-{\it small}, which generates single irrelevant words instead of empty sequences.\footnote{We report these results to \Cref{app:speech_output_type}.}

\section{Conclusion}
Our work contributes to a growing body of research on the interpretability of language models. By introducing the DecoderLens, we provide insights into how intermediate encoder representations of encoder-decoder Transformers influence decoder predictions. 
We apply our method to various models and tasks, finding that intermediate outputs can provide valuable insights into the model's decision-making process. 
In particular, our findings reveal how ``simpler'' subtasks (e.g., simple conjunctive logic formulas, high-resource MT, speech transcription) are captured by early encoder layers with high precision and persist up to the final model output through the residual stream, while more challenging tasks (e.g., complex logic formulas, low-resource MT, speech translation) are only addressed by final encoder layers.
{We also find some evidence that early layer outputs are more {\it local} solutions. 
Errors in variable assignments in the middle layer of the Transformer trained on propositional logic are due to the model solving subparts of the sentence independently.
Additionally, translations from earlier layers of NLLB adhere more closely to the word order of the input. %

These observations are} in line with previous work on probing, which showed that linguistic subtasks in LMs are performed at different stages in Transformers \citep{tenney_bertrediscoversclassical_2019, peters-etal-2018-dissecting}. Moreover, it provides evidence that model predictions are \textit{refined iteratively} also across encoder layers, complementing previous work on decoder-only models. 
By verbalizing the knowledge encoded in intermediate model layers, DecoderLens can provide useful and human-interpretable insights into the evolution of model predictions, complementing other interpretability techniques for the study of neural language models.

Future work could explore the application of DecoderLens to the Universal Transformer~\citep{dehghani2018universal}, especially for algorithmic tasks~\citep{csordas_devildetailsimple_2022a} where its intermediate representation might be more interpretable and compositional thanks to weight sharing. 
Moreover, the tendency of earlier layers to produce simpler generations can be connected to outputs produced during early stages of model training~\citep{voita-etal-2021-language}. 
In this context, DecoderLens might be used to investigate the relation between training dynamics and information geometry across model layers~\citep{DBLP:conf/acl/ChoshenHWA22,tunedlens}.
{Lastly, DecoderLens could be used as diagnostic tool to investigate where wrong model predictions emerge, which is useful for both interpretability purposes and model improvement through early exiting strategies.}

\section{Limitations}

One important concern regarding the direct use of intermediate representations to make predictions is that of {\it representational drift}: features may be represented differently in earlier encoder layers, reducing the ability of the decoder to use this information. This manifested in particular  in the form of hallucinated or empty DecoderLens predictions for early encoder layers.
While this representational misalignment could be mitigated by {\it tuning} representations to match the space of final layers~\citep{tunedlens, yomdin-etal-2023-jump}, we limit our analysis to the direct application of DecoderLens without any additional training.

We note that DecoderLens does not reveal \emph{where} within a layer a specific subtask is solved (i.e., which heads or MLP-units within the layer are responsible), nor does it reveal \emph{how} subtasks are solved. For this reason, while we consider our method promising to provide a more intuitive overview of encoder capabilities, we also believe it should be complemented with other approaches to obtain fine-grained insights into model predictions.

Finally, although our experiments span several encoder-decoder models and tasks, our evaluation is limited to small model sizes (<700M parameters) due to limited computational resources. It remains to assess whether our findings using the DecoderLens method still apply to larger models.

\paragraph*{Acknowledgments}
We gratefully acknowledge support for this research from the Institute for Logic, Language and Computation at the University of Amsterdam (JJ, WZ), and from the Netherlands Organization for Scientific Research (NWO), through the NWA-ORC grant NWA.1292.19.399 for `InDeep' (supporting HM, GS), and through the Graviation grant 024.001.006 for `Language in Interaction' (supporting AL). We also thank SURF and the Faculty of Science (FNWI, UvA) for the use of the National Supercomputer Snellius.


\bibliography{anthology,custom}
\bibliographystyle{acl_natbib}

\appendix

\newpage\section{Propositional Logic}\label{sec:logic:appendix}
\begin{figure}[!h]
    \centering
    \includegraphics[width=\linewidth]{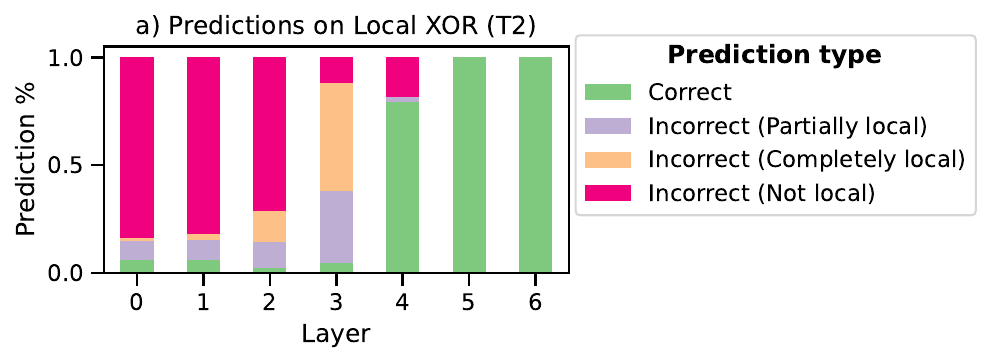}
    \includegraphics[width=\linewidth]{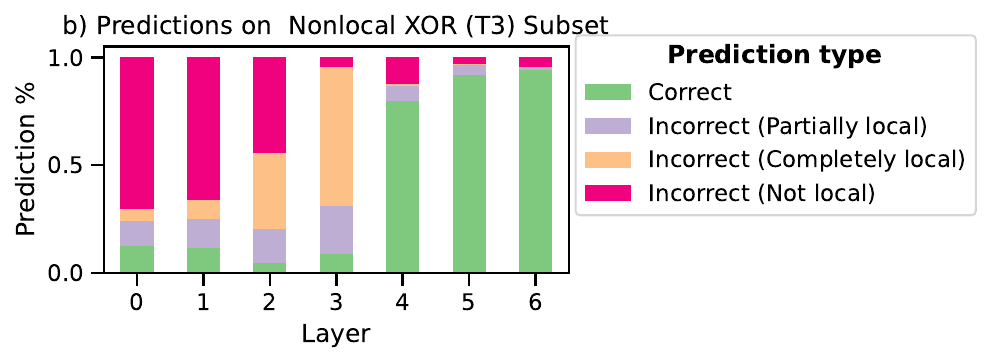}
    \includegraphics[width=\linewidth]{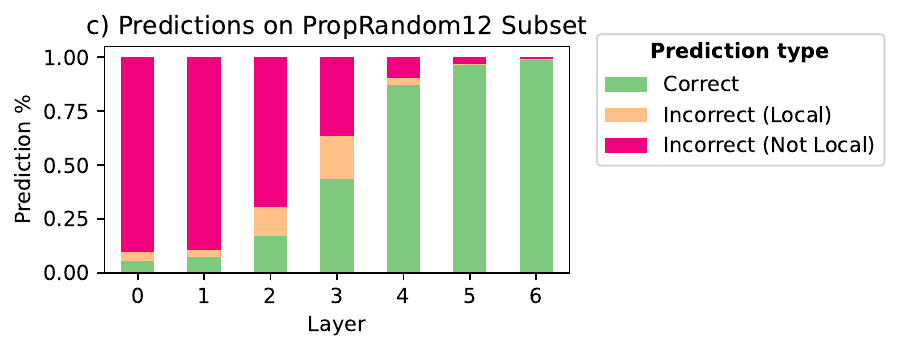}
    \caption{Distribution of the types of predictions on three small datasets. A {\it local} solution means the layer assigns {\it false} (\texttt{0}) to a variable if it occurs in the input negated, and {\it true} (\texttt{1}) if the variable appears non-negated.
    We therefore consider only the subset of data for which each variable either only occurs negated or only occurs non-negated.
    Layer 3 produces the largest number of local solutions in all cases.
    }
    \label{fig:logic_appendix}
\end{figure}

\section{Machine Translation}
\subsection{Additional Examples of DecoderLens Translations}
\label{app:mt_examples}

Tables~\ref{tab:mt_examples_1} and~\ref{tab:mt_examples_2} showcase some additional examples for some of the selected translation directions.

\begin{table*}
\centering
\footnotesize
\begin{tabular}{p{\textwidth}}
\toprule
\textbf{Source:}     Dr. Ehud Ur, professor of medicine at Dalhousie University in Halifax, Nova Scotia and chair of the clinical and scientific division of the Canadian Diabetes Association cautioned that the research is still in its early days. \\
\textbf{Reference:}  Le Dr Ehud Ur, professeur de médecine à l'Université Dalhousie de Halifax (Nouvelle-Écosse) et président de la division clinique et scientifique de l'Association canadienne du diabète, a averti que la recherche en était encore à ses débuts. \\
\midrule
\textbf{L1:}         Le professeur de la médecine, le professeur de la médecine, le professeur de la médecine, le professeur de la médecine, le professeur de la médecine, le professeur de la médecine, le professeur de la médecine, le professeur de la médecine, le [...] \\
\textbf{L2:}         Le Dr. Ehud, le professeur de la médecine, a déclaré: "La recherche de la médecine est une expérience de la médecine de la médecine, mais je suis en train de me dire que je suis en train de me lancer dans la recherche. \\
\textbf{L3:}         Le professeur de la médecine de l'Université de Halifax et de la division scientifique de l'Association canadienne de la recherche est toujours dans la recherche de la recherche de la recherche de [...] \\
\textbf{L4:}         Le Dr. Ehud, professeur de l'Université de Halifax, a présenté la recherche de la division scientifique de l'Académie canadienne de la recherche et de la recherche. \\
\textbf{L5:}         Le Dr. Ehud, professeur de médecine à l'Université de Halifax, et le président de la division scientifique du Diabetes Association canadien, ont fait état de la recherche qui se déroule dans ses premières années. \\
\textbf{L6:}         Le professeur de médecine de l'Université de Halifax, le professeur d'Eud Ur, et le président de la division scientifique du Diabète canadien, ont fait remarquer que la recherche est toujours en cours. \\
\textbf{L7:}         Le professeur de médecine Ehud Ur, professeur de médecine à l'Université de Halifax, en Nouvelle-Écosse, et président de la division clinique et scientifique de l'Association canadienne du Diabète a mis en garde que la recherche est toujours dans ses premiers jours. \\
\textbf{L8:}         Le professeur de médecine de l'Université de Dalhousie, en Nouvelle-Écosse, et président de la division clinique et scientifique de l'Association canadienne du diabète, a souligné que la recherche est encore à ses débuts. \\
\textbf{L9:}         Le professeur de médecine de l'Université de Dalhousie, en Nouvelle-Écosse, et président de la division clinique et scientifique de l'Association canadienne du diabète, Dr. Ehud Ur, a souligné que la recherche est encore en début de phase. \\
\textbf{L10:}        Le Dr Ehud Ur, professeur de médecine à l'Université Dalhousie à Halifax, en Nouvelle-Écosse, et président de la division clinique et scientifique de l'Association canadienne du diabète, a averti que la recherche est encore dans ses premiers jours. \\
\textbf{L11:}        Le professeur de médecine de l'université Dalhousie à Halifax, en Nouvelle-Écosse, et président de la division clinique et scientifique de l'Association canadienne du diabète, Dr Ehud Ur, a averti que la recherche était encore à ses débuts. \\
\textbf{L12:}        Le Dr Ehud Ur, professeur de médecine à l'Université Dalhousie à Halifax, en Nouvelle-Écosse, et président de la division clinique et scientifique de l'Association canadienne du diabète, a averti que la recherche est encore à ses débuts. \\
\bottomrule
\bottomrule
\textbf{Source:} "We now have 4-month-old mice that are non-diabetic that used to be diabetic," he added. \\
\textbf{Reference:} "Abbiamo topi di quattro mesi che prima erano diabetici e ora non lo sono più", ha aggiunto. \\
\midrule
\textbf{L1:} "Ci sono due problemi che hanno portato a questo problema, ma non ci sono problemi che possono essere risolti. \\
\textbf{L2:} "Abbiamo 4-month-diabetic che sono utilizzati per essere, che sono utilizzati per il diabete. \\
\textbf{L3:} "Abbiamo 4-month-that sono i non-diabetic che sono utilizzati, che sono aggiunti". \\
\textbf{L4:} "Abbiamo ora 4 mesi che sono i non-diabetic che sono utilizzati per essere diabetico," ha aggiunto. \\
\textbf{L5:} "Abbiamo ora 4 mesi di cicli che sono non-diabetic che hanno usato per essere diabetico, "ha aggiunto. \\
\textbf{L6:} "Abbiamo ora 4 mesi di topi che sono non-diabetico che hanno usato per essere diabetico", ha aggiunto. \\
\textbf{L7:} "Abbiamo ora topi di 4 mesi che sono non diabetici che erano diabetici", aggiunge. \\
\textbf{L8:} "Abbiamo ora topi di 4 mesi che non sono diabetici e che erano diabetici", aggiunge. \\
\textbf{L9:} "Abbiamo ora i topi di 4 mesi che non sono diabetici e che erano diabetici", ha aggiunto. \\
\textbf{L10:} "Abbiamo ora topi di 4 mesi che non hanno il diabete e che erano diabetici", ha aggiunto. \\
\textbf{L11:} "Ora abbiamo topi di 4 mesi che non hanno il diabete e che erano diabetici", ha aggiunto. \\
\textbf{L12:} "Ora abbiamo topi non diabetici di 4 mesi che erano diabetici", ha aggiunto. \\
\bottomrule
\bottomrule
\textbf{Source:} Plus de 4 000 000 personnes sont se sont rendues à Rome pour l'enterrement.\\
\textbf{Reference:} Over four million people went to Rome to attend the funeral.\\
\midrule
\textbf{L1:} 4 people are in the process of getting their hands on the car. \\
\textbf{L2:} 4 000 people are in the city. \\
\textbf{L3:} More than 4 000 people are being sent to Rome for their own country. \\
\textbf{L4:} More than 4 000 people are being made to Rome for the entertainment. \\
\textbf{L5:} More than 4 000 people have been to Rome for the entertainment. \\
\textbf{L6:} More than 4 000 000 people have gone to Rome for the funeral. \\
\textbf{L7:} More than 4,000,000 people have gone to Rome for the funeral. \\
\textbf{L8:} More than 4 000 000 people have gone to Rome for the funeral.  \\
\textbf{L9:} More than 4,000,000 people have come to Rome for the funeral.  \\
\textbf{L10:} More than 4 million people attended the funeral in Rome. \\
\textbf{L11:} More than four million people have come to Rome for the funeral. \\
\textbf{L12:} More than four million people went to Rome for the funeral. \\
\bottomrule
\end{tabular}
\caption{Examples for English $\rightarrow$ French, English $\rightarrow$ Italian and French $\rightarrow$ English translation using DecoderLens on NLLB.}
\label{tab:mt_examples_1}
\end{table*}

\begin{table*}
\centering
\footnotesize
\begin{tabular}{p{\textwidth}}
\toprule
\textbf{Source:}     While one experimental vaccine appears able to reduce Ebola mortality, up until now, no drugs have been clearly demonstrated suitable for treating existing infection. \\
\textbf{Reference:}  Eén experimenteel vaccin lijkt in staat te zijn de ebola-sterfte terug te dringen, maar tot nu toe zijn nog geen medicijnen duidelijk geschikt voor de behandeling van bestaande infecties. \\
\midrule
\textbf{L1:}         Een vaccinatie is een goede manier om de ziekte te voorkomen. \\
\textbf{L2:}         Een Ebola-infectie is een gevaarlijk risico. Het is een gevaarlijk risico dat de ziekte van de ziekte van de ziekte van de ziekte van de ziekte van de ziekte kan voorkomen. \\
\textbf{L3:}         Terwijl de Ebola-vaccinatie wordt verminderd, is de aanwezigheid van een Ebola-vaccinatie niet mogelijk. \\
\textbf{L4:}         Hoewel de ebola-vaccinatie in de praktijk wordt beperkt, wordt de ebola-vaccinatie niet meer gebruikt. \\
\textbf{L5:}         Terwijl een experimentele vaccine lijkt te verminderen Ebola-taligheid, is er tot nu toe geen drugs die geschikt zijn voor het behandelen van bestaande infectie. \\
\textbf{L6:}         Terwijl een experimentele vaccine de Ebola-sterfte kan verminderen, zijn er tot nu toe geen geneesmiddelen die geschikt zijn voor de behandeling van bestaande infectie. \\
\textbf{L7:}         Hoewel een experimentele vaccine de Ebola-sterfte kan verminderen, is er tot nu toe geen enkele geneesmiddel die geschikt is voor de behandeling van bestaande infectie. \\
\textbf{L8:}         Hoewel één experimentele vaccine de Ebola-sterfte kan verminderen, is er tot nu toe geen enkele geneesmiddel geschikt voor de behandeling van bestaande infectie. \\
\textbf{L9:}         Hoewel een experimental vaccin de sterfte van Ebola kan verminderen, is er tot nu toe geen enkel geneesmiddel geschikt voor de behandeling van bestaande infecties. \\
\textbf{L10:}        Hoewel een experimentele vaccine de sterfte van Ebola lijkt te verminderen, is tot nu toe geen enkele geneesmiddel duidelijk geschikt voor de behandeling van bestaande infectie. \\
\textbf{L11:}        Hoewel één proefvaccin de sterfte van Ebola lijkt te verminderen, is tot nu toe geen enkel geneesmiddel duidelijk aangetoond dat het geschikt is voor de behandeling van bestaande infectie. \\
\textbf{L12:}        Hoewel één experimentele vaccin de sterfte van ebola lijkt te kunnen verminderen, is tot nu toe geen enkel geneesmiddel duidelijk aangetoond dat geschikt is voor de behandeling van bestaande infectie. \\
\bottomrule
\bottomrule
\textbf{Source:}     Volgens wetenschappers was het verenkleed van dit dier kastanjebruin met een bleke of carotenoïdekleurige onderzijde. \\
\textbf{Reference:}  Scientists say this animal's plumage was chestnut-brown on top with a pale or carotenoid-colored underside. \\
\midrule
\textbf{L1:}         According to the Bible, the dead were not born, and the dead were not born, and   [...] the dead were not yet alive. \\
\textbf{L2:}         According to the Bible, the animal was not a good animal, but a good animal. \\
\textbf{L3:}         According to the scientists, this was a very dangerous disease. \\
\textbf{L4:}         According to the scientists, this was a kind of animal that was not a carotenoid. \\
\textbf{L5:}        ..................................................................................................................................................................................................... \\
\textbf{L6:}         According to scientists, the crest of this animal was a brown or carotenoid-coloured crest. \\
\textbf{L7:}         According to scientists, the embroidery of this animal was chestnut with a pale or carotenoid-coloured underside. \\
\textbf{L8:}         According to scientists, the animal was a brownish-brown animal with a pale or carotenoid undercoat. \\
\textbf{L9:}         According to scientists, the animal was a brownish-brown, with a pale or carotenoid undercoat. \\
\textbf{L10:}        According to scientists, the animal's undercoat was brown with a pale or carotenoid underside. \\
\textbf{L11:}        According to scientists, the animal's embroidery was chestnut with a pale or carotenoid undercoat. \\
\textbf{L12:}        Scientists say the animal's disguise was chestnut brown with a pale or carotenoid undercoat. \\
\bottomrule
\bottomrule
\textbf{Source:} L'annuncio è stato fatto a seguito di un colloquio telefonico tra Trump e il presidente turco Recep Tayyip Erdoğan. \\
\textbf{Reference:} The announcement was made after Trump had a phone conversation with Turkish President Recep Tayyip Erdoğan. \\
\midrule
\textbf{L1:} A phone call from the president of the United States of America was made. \\
\textbf{L2:} The president's speech was broadcast on the Internet. \\
\textbf{L3:} The president of the Republic of Turkey, President Tayyip Erdogan, is a member of the Turkish parliament. \\
\textbf{L4:} The announcement was made at a meeting of the President of the Republic of Turkey, President of the Republic of Turkey, and the President of the [...] \\
\textbf{L5:} The announcement was made following a phone call between the President of Turkey, President Tayyip Erdogan. \\
\textbf{L6:} The announcement was made following a phone call between Trump and the Turkish President, Recep Tayyip Erdoğan. \\
\textbf{L7:} The announcement was made following a phone conversation between Trump and the Turkish President Recep Tayyip Erdoğan. \\
\textbf{L8:} The announcement was made following a phone conversation between Trump and Turkish President Recep Tayyip Erdoğan. \\
\textbf{L9:} The announcement was made following a phone conversation between Trump and Turkish President Recep Tayyip Erdoğan. \\
\textbf{L10:} The announcement was made following a phone conversation between Trump and Turkish President Recep Tayyip Erdoğan. \\
\textbf{L11:} The announcement was made following a phone conversation between Trump and Turkish President Recep Tayyip Erdoğan. \\
\textbf{L12:} The announcement was made following a phone conversation between Trump and Turkish President Recep Tayyip Erdoğan. \\
\bottomrule
\end{tabular}
\caption{Examples for English $\rightarrow$ Dutch, Dutch $\rightarrow$ English and Italian $\rightarrow$ English translation using DecoderLens on NLLB.}
\label{tab:mt_examples_2}
\end{table*}

\clearpage
\section{Speech to Text}
\subsection{WER results for other model sizes}
\label{app:speech_wer}

\begin{figure}[h!]
    \centering
    \includegraphics[width=\linewidth]{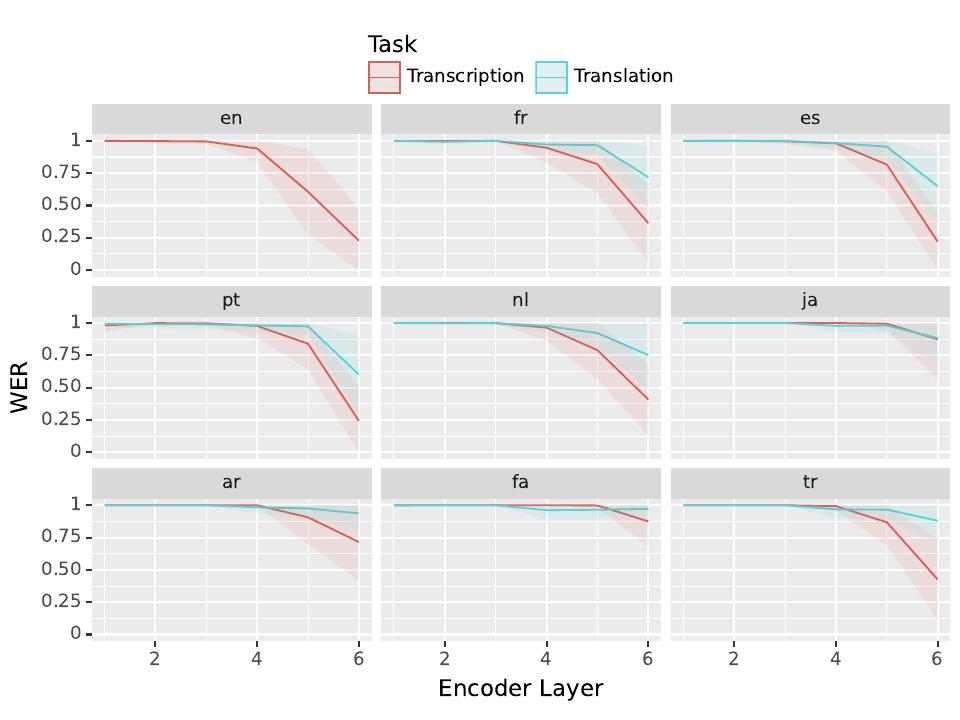}
    \caption{The change in Word Error Rate (wer) of Whisper-base for transcription and translation, averaged over our test examples, w.r.t number of encoder layer used at inference. Shaded areas show std.}
\end{figure}

\begin{figure}[h!]
    \centering
    \includegraphics[width=\linewidth]{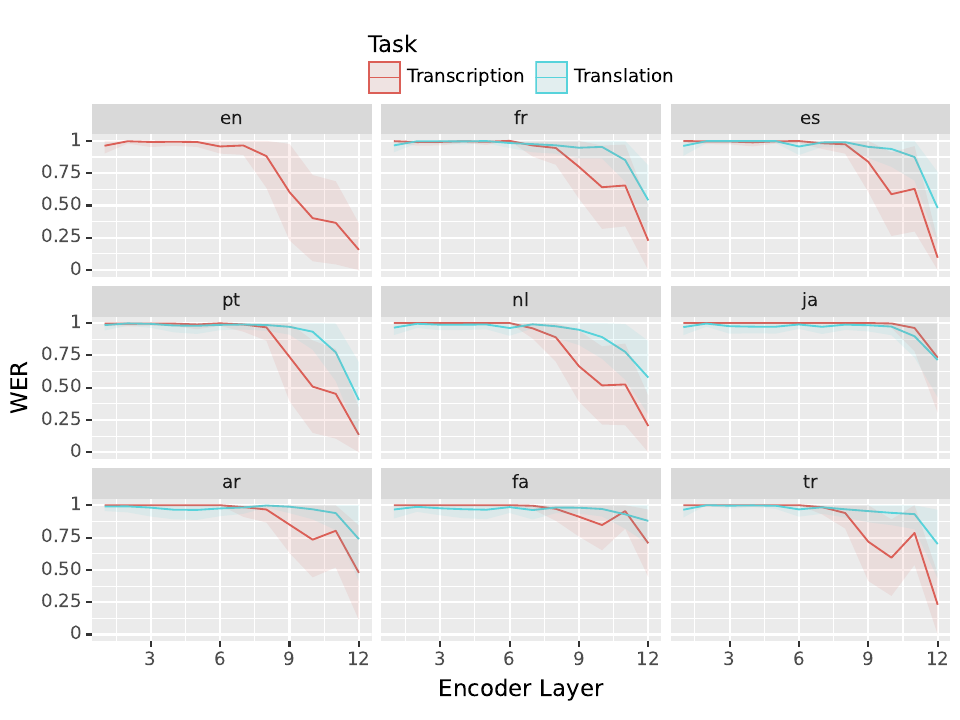}
    \caption{The change in Word Error Rate (wer) of Whisper-small for transcription and translation, averaged over our test examples, w.r.t number of encoder layer used at inference. Shaded areas show std.}
\end{figure}

\subsection{Distribution of output types for other model sizes}
\label{app:speech_output_type}

\begin{figure}[h!]
    \centering
    \includegraphics[width=\linewidth]{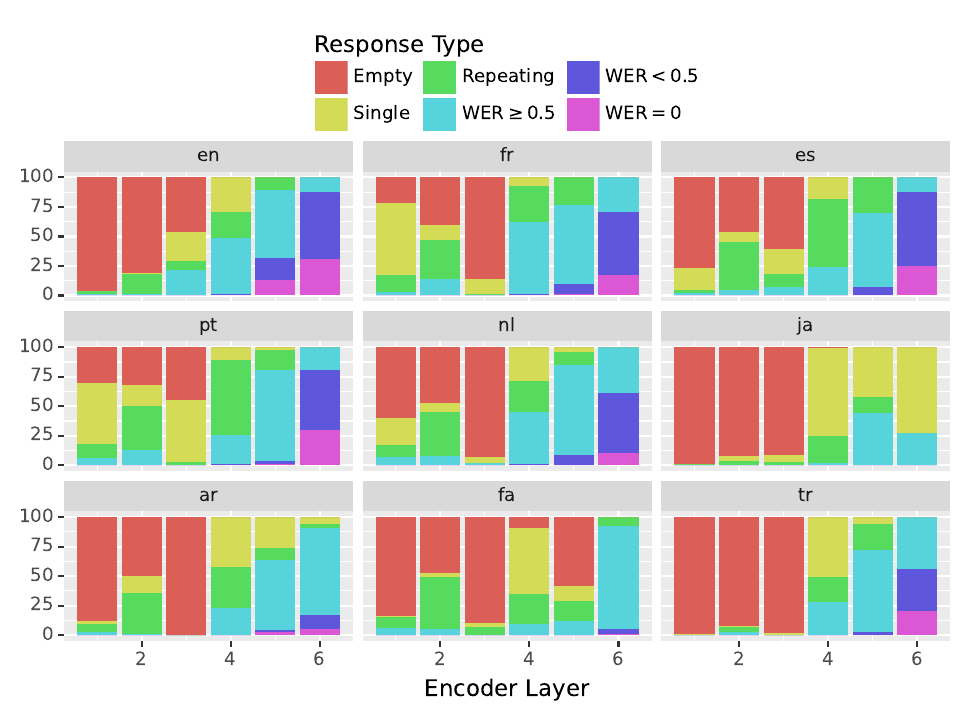}
    \caption{Distribution of Whisper-{\it base} output types when transcribing w.r.t number of encoder layer used at inference.}
\end{figure}

\begin{figure}[h!]
    \centering
    \includegraphics[width=\linewidth]{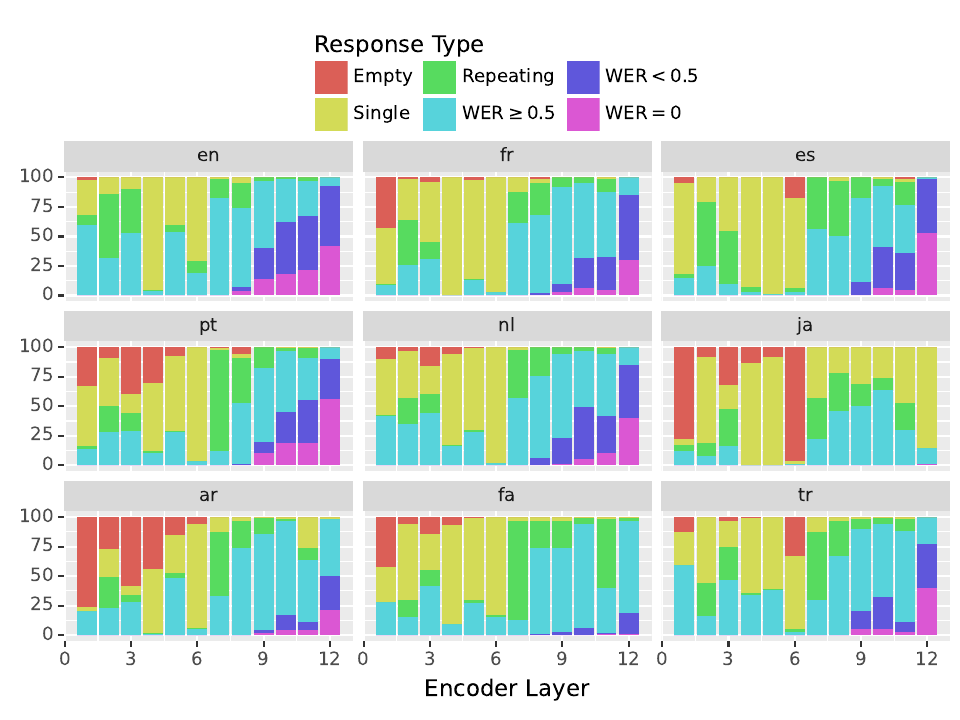}
    \caption{Distribution of Whisper-{\it small} output types when transcribing w.r.t number of encoder layer used at inference.}
\end{figure}

\begin{figure}[h!]
    \centering
    \includegraphics[width=\linewidth]{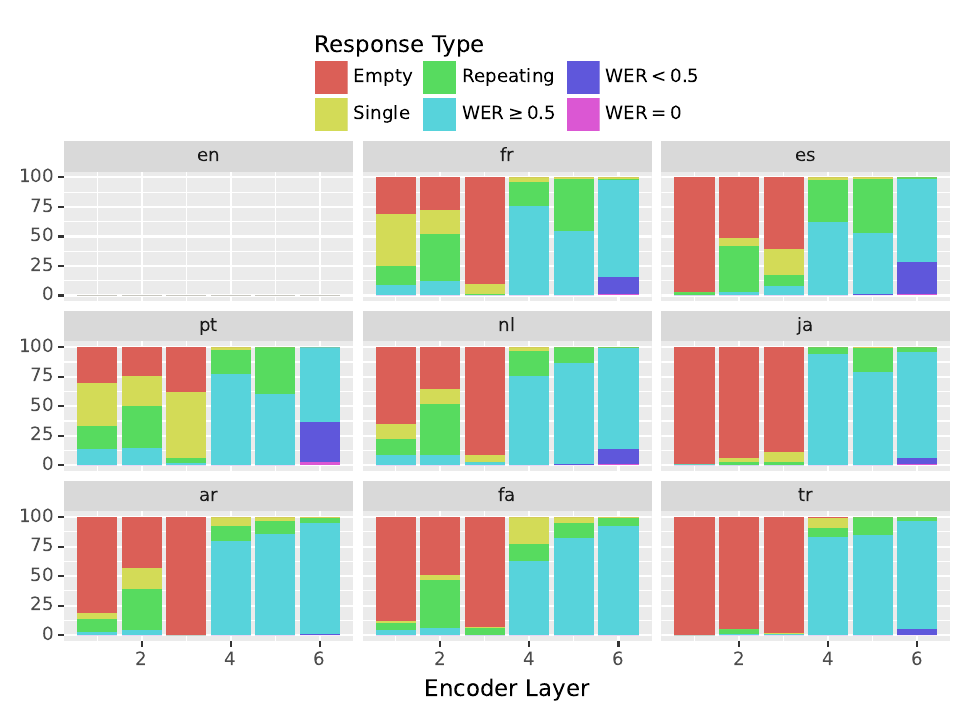}
    \caption{Distribution of Whisper-base output types when translating to English w.r.t number of encoder layer used at inference.}
\end{figure}

\begin{figure}[h!]
    \centering
    \includegraphics[width=\linewidth]{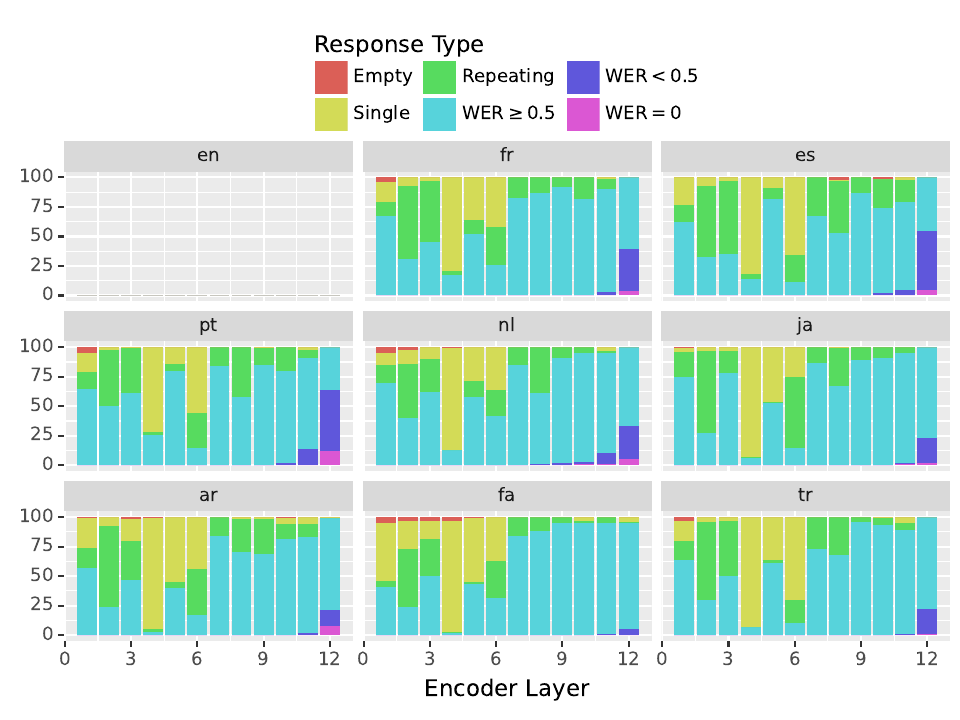}
    \caption{Distribution of Whisper-small output types when translating to English w.r.t number of encoder layer used at inference.}
\end{figure}

\begin{figure}[h!]
    \centering
    \includegraphics[width=\linewidth]{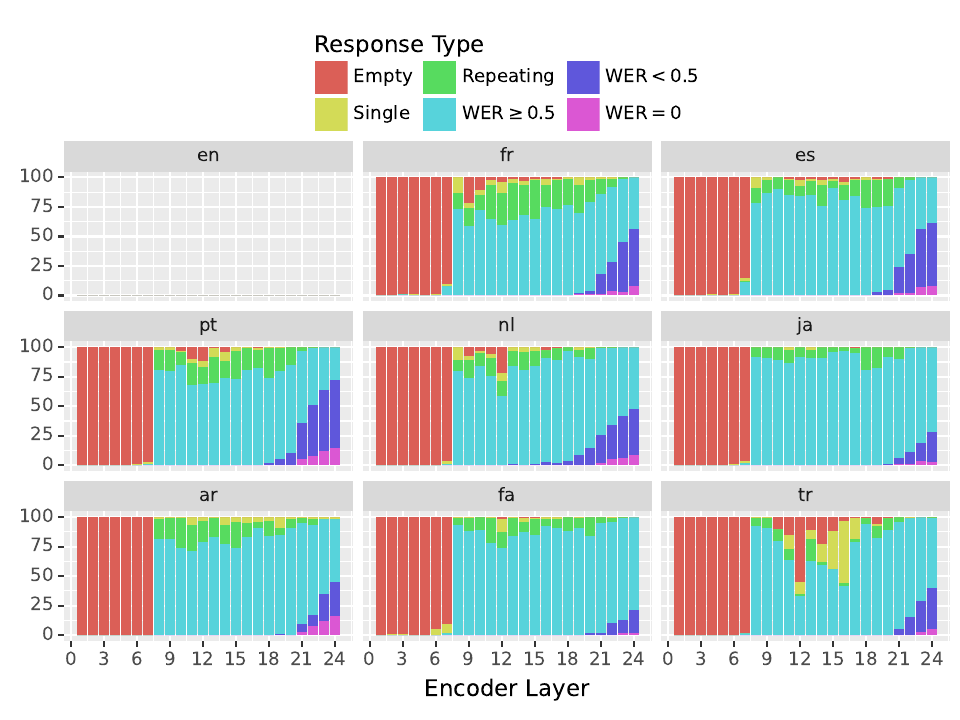}
    \caption{Distribution of Whisper-medium output types when translating to English w.r.t number of encoder layer used at inference.}
\end{figure}

\end{document}